\documentclass[11pt]{article}

\usepackage[a4paper,margin=1in]{geometry}
\usepackage[T1]{fontenc}
\usepackage{lmodern}
\usepackage{microtype}
\usepackage{setspace}
\usepackage{amsmath,amssymb}
\usepackage{booktabs}
\usepackage{graphicx}
\usepackage{caption}
\usepackage{subcaption}
\usepackage{array}
\usepackage{multirow}

\usepackage[round,authoryear]{natbib}
\usepackage[colorlinks=true,linkcolor=black,citecolor=blue,urlcolor=blue]{hyperref}

\title{\vspace{-2em}AI Receptivity or AI Adoption Breadth?\\
A Tool-Specific Reanalysis of the\\ Lower-Literacy/Higher-Usage Link}
\author{Hristo Inouzhe Valdes\\
\normalsize Associate Professor\\
\normalsize Departamento de Matemáticas, Universidad Aut\'onoma de Madrid\\
\normalsize \href{mailto:hristo.inouze@uam.es}{hristo.inouze@uam.es}}
\date{\today}

\begin{document}
\maketitle

\begin{abstract}
\noindent
Recent evidence reported by \citet{tully2025lower} suggests that lower artificial
intelligence (AI) literacy predicts greater receptivity toward AI. We revisit this
claim using the public data from Study~3 of that article, which measures past
usage of five AI tool categories on a five-point frequency scale. We first
reproduce the negative association between AI literacy and aggregate AI usage
using OLS on participant-level averages, binary logit, ordered logit, and
multinomial logit specifications. We then show that the aggregate relationship
masks substantial heterogeneity by tool type. In our demographic-adjusted
primary specification, AI literacy does not significantly predict text AI usage
(ordered-logit $\beta = -0.090$, $p = .387$), whereas it remains a strong
predictor of non-text AI adoption ($\beta = -0.377$, $p < .001$). The non-text
effect is also robust under Tully et al.'s original Study~3 control
specification
($\beta = -0.502$, $p < .001$). Binary, ordered-logit, and multinomial
specifications suggest that the non-text relationship is primarily an
adoption/non-adoption pattern rather than evidence of intensive use: the
demographic-adjusted odds ratio of ever having used a non-text AI tool is
$0.68$. Thus, in the study that measures self reported past usage rather than stated
preferences, the evidence does not support a simple claim that lower AI literacy
predicts greater receptivity to AI in general. It points instead to a narrower
pattern of broader adoption across lower-penetration, non-text AI tools.

\end{abstract}

\section{Introduction}

The proliferation of generative AI products has renewed marketing scholars'
interest in the individual-level antecedents of AI adoption
\citep{puntoni2021consumers,hermann2024aiconsumer}. Among recent contributions,
\citet{tully2025lower} document a counterintuitive lower-literacy, higher-receptivity
link across seven studies: consumers who score lower on objective tests of AI
knowledge are, on average, more receptive to AI. Study~3 is especially useful
for reanalysis because, unlike Studies~4
and~6, it uses reported past usage of real AI products, rather than stated
preferences for hypothetical tasks, as the dependent variable. This makes it a
more conservative test of the proposed relationship and, accordingly, a key
target for replication and robustness checks.

The original Study~3 (N$=$401, Amazon Mechanical Turk) measures usage frequency
over the previous six months for five AI tool categories: digital image
generators (e.g., DALL-E), AI-powered productivity tools (e.g., Zapier),
AI-powered design services (e.g., Canva), AI-powered health apps (e.g.,
Headspace), and AI-powered writing assistants (e.g., ChatGPT). Each tool is
rated on a five-point frequency scale ranging from Never to
Weekly. The authors average these five items into an AI receptivity
index and regress the index on AI literacy.

We see two reasons to revisit this specification. First, averaging ordinal
five-point items into a single continuous index is a standard but contested
practice, and the field has accumulated substantial methodological evidence
that doing so can distort estimates of treatment effects and even invert their
sign \citep{liddell2018analyzing,agresti2010analysis}. Second, and more
substantively, the five tool categories arguably tap heterogeneous behavioral
constructs. Text AI (writing assistants) had become highly visible after the
release of ChatGPT, whereas image generators, productivity bots, website
builders, and meditation apps were plausibly less familiar products.
Aggregating them
into a single ``AI receptivity'' index conflates intensive text AI use with
broader adoption of AI-labelled consumer tools.

In this paper we revisit Study~3 along both dimensions. We do not dispute the
aggregate lower-literacy/higher-usage association; rather, we show that its
substantive meaning changes when AI usage is decomposed by tool type and
modelled as ordinal adoption data. The reanalysis makes three points:

\begin{enumerate}
  \item Replication: the pooled Study~3 association is real. We
  reproduce the original direction under OLS, binary logit, ordered logit,
  and multinomial logit.
  \item Measurement: averaging ordinal usage responses across
  heterogeneous AI categories compresses different behavioral processes into a
  single index.
  \item Interpretation: the relationship is much stronger and more robust for
  non-text AI tools than for text AI alone, and the non-text effect is
  concentrated at the adoption margin. Study~3 therefore does not support the
  broadest reading of the original claim as a general lower-literacy,
  higher-AI-receptivity effect. It is better interpreted as evidence about
  non-text AI adoption breadth.
\end{enumerate}

\section{Background}

\paragraph{Algorithm aversion and algorithm appreciation.} A long literature in
marketing and judgment-and-decision-making documents systematic individual
responses to algorithmic agents. \citet{dietvorst2015algorithm} show that
people abandon algorithms after observing errors, even when the algorithm
outperforms a human alternative. \citet{logg2019algorithm} show the
complementary phenomenon, algorithm appreciation, in numerical estimation
tasks. \citet{castelo2019taskdependent} and \citet{longoni2022wordofmachine}
qualify both patterns by showing that task type (objective vs.\ subjective;
utilitarian vs.\ hedonic) moderates consumer responses, while
\citet{longoni2019resistance} document resistance to medical AI driven by
uniqueness neglect. \citet{yalcin2022thumbs} show that the direction of the
algorithmic decision (favorable vs.\ unfavorable) matters for consumer
reactions, and \citet{debellis2023meaning} show that the perceived meaning of
manual labor impedes the adoption of autonomous products. Across this
literature, contextual and task-related factors play a central role, and
individual-level determinants of receptivity remain comparatively
understudied.

\paragraph{AI literacy.} The construct of AI literacy itself was articulated
by \citet{long2020aileiteracy} as a set of competencies needed to evaluate,
communicate with, and use AI in everyday contexts. Subsequent reviews
\citep{ng2021conceptualizing} adapted classical literacy frameworks to
characterize AI-specific knowledge, use, evaluation, and ethics. Against this
backdrop, \citet{tully2025lower} make the novel claim that higher AI literacy
decreases receptivity to AI, because demystifying AI dispels
perceptions of magic and the awe such perceptions evoke.

\paragraph{Tool-level heterogeneity in AI adoption.} Recent macro evidence
underscores why aggregating AI tool categories is risky. \citet{mcelheran2024ai}
document that, as of 2018, fewer than 6\% of U.S.\ firms used any of five core
AI technologies, and adoption was strongly concentrated by industry and firm
size. \citet{brynjolfsson2025generative} show that productivity gains from
generative AI assistants are heterogeneous across workers, with the largest
gains accruing to lower-skill agents. \citet{acemoglu2020robots} make a
parallel point for industrial robots. These results suggest that
``AI adoption'' is not a single behavioral phenomenon but a family of
tool-specific decisions whose drivers may differ across tool categories.

\paragraph{Ordinal versus metric analysis.} On the methodological side,
\citet{mccullagh1980regression} introduced the proportional-odds model that we
adopt here, and \citet{agresti2010analysis} provides a comprehensive treatment
of ordinal regression. \citet{liddell2018analyzing} review hundreds of
psychology articles and document that treating ordinal Likert responses as
metric can yield false positives, false negatives, and, in extreme
cases, sign inversions of estimated effects. Their recommendation, which we
follow, is to estimate ordered-probit or ordered-logit models on the
item-level data rather than OLS on averaged indices.

\section{Original Study~3 and Analytic Concern}

\citet{tully2025lower} report that lower AI literacy was associated with
greater frequency of AI usage
($B = -0.09$, $\mathrm{SE} = .02$, $t(399) = -5.73$, $p < .001$), with the
coefficient remaining significant after controlling for technology readiness,
general knowledge, motivation for autonomy, and gender ($B = -0.11$, $p <
.001$). The dependent variable is the mean of five items
$Y_{ij} \in \{1, 2, 3, 4, 5\}$ measuring usage frequency of image generators,
productivity tools, website builders, health apps, and writing assistants.

Two features of the design motivate our reanalysis. First, the five items
correspond to qualitatively different AI products. Their empirical
distributions differ sharply: text AI is the only item with substantial support
across the full five-point scale, with at least 50 respondents in every response
category and roughly two-thirds of respondents reporting at least some use. By
contrast, the four non-text categories are heavily concentrated at the lowest
category (``Never''), with 65--78\% of respondents reporting no usage in the
previous six months and thinly populated upper categories. Second, the
outcome
\[
Y_{ij} \in \{1, 2, 3, 4, 5\}
\]
is ordinal, sparse for several tools, and repeated within participant. This
matters because the text AI item is the best-sampled ordinal test of whether
lower AI literacy predicts higher usage across the frequency scale, whereas the
non-text items primarily distinguish non-users from everyone else. Averaging is
not inherently invalid, but it conflates two distinct margins:
(i) \emph{adoption} (whether respondents use a tool at all) and
(ii) \emph{intensity} (how often they use it conditional on adoption). When
those margins differ across tools, the pooled effect estimate need not have a
clean substantive interpretation.

\section{Reanalysis Strategy}

We download the public Study~3 data from
\href{https://researchbox.org/1491}{ResearchBox \#1491} and use the AI
literacy score (\texttt{SC0}, summed correct answers on the 17-item
AI-constructed measure) and the five usage items. Our primary specification is
demographic-adjusted: age, income, general knowledge, motivation for autonomy,
and a male-gender indicator. We use this specification because it retains the
full Study~3 analytic sample (N$=$401) while adjusting for demographics and the
two non-technology individual-difference covariates emphasized in the original
paper. As a robustness check, we re-estimate the key models using the control
specification for the original Study~3 regression results reported in
Table~4 of \citet{tully2025lower}: technology readiness, general knowledge,
motivation for autonomy, and a male-gender indicator (N$=$379 because
technology readiness has missing observations). We standardize the AI literacy score and continuous
covariates so that coefficients are interpretable as the effect of a
one-standard-deviation increase. We then fit four model families across three
outcome groupings.

\paragraph{Model families.}
\begin{enumerate}
  \item OLS on the participant-level average. This is the closest
    match to the original Study~3 specification and serves as our pooled
    benchmark.
  \item Binary logit on item-level data with task fixed effects. We
    consider two thresholds: $Y > 3$ (``frequent use'') and $Y > 1$
    (``adoption'').
  \item Ordered logit (proportional-odds model)
    \citep{mccullagh1980regression}. This is the natural model for an
    ordered Likert response with $K = 5$ categories and is robust to violation
    of metric-scale assumptions \citep{liddell2018analyzing}.
  \item Multinomial logit, treating the five categories as nominal.
    This is a stress test: it does not impose proportional odds or any
    ordering and is therefore robust to violations of either.
\end{enumerate}

\paragraph{Outcome groupings.}
\begin{enumerate}
  \item Pooled (5 tools): the original Study~3 outcome.
  \item Text only: \texttt{AI\_text} alone.
  \item Non-text only: \texttt{AI\_image},
    \texttt{AI\_productivity}, \texttt{AI\_website}, \texttt{AI\_healthapp}.
\end{enumerate}

\paragraph{Implementation.} All models are fit in Python with
\texttt{statsmodels}. OLS and binary-logit standard errors are
heteroskedasticity-robust (\texttt{HC3}). Item-level models include
task fixed effects. Full code, a Jupyter notebook, and pre-generated figures
are provided as supplementary material.

\section{Results}

\subsection{Replication of the pooled effect}

The demographic-adjusted pooled five-tool association is negative and
statistically significant in every specification. OLS on the participant-level
average gives $\hat\beta = -0.181$ ($p = .001$), after standardization. The
ordered logit gives $\hat\beta = -0.307$ ($p < .001$), the binary logit at the
scale midpoint gives $\hat\beta = -0.320$ ($p < .001$, odds ratio $=0.73$), and
the binary adoption logit gives $\hat\beta = -0.330$ ($p < .001$, odds ratio
$=0.72$). The multinomial logit yields the same negative direction across all
four non-reference categories. We therefore reproduce the original pooled
lower-literacy/higher-usage association before decomposing it by tool type.

\subsection{Text vs.\ non-text decomposition}

Table~\ref{tab:decomposition} re-estimates the ordered-logit and binary-logit
specifications separately on the text AI outcome and on the four non-text
outcomes. The demographic-adjusted contrast is clear. For text AI alone, the
ordered-logit coefficient is small and not significantly different from zero
($\hat\beta = -0.090$, $\mathrm{SE} = 0.104$, $p = .387$); the binary
``frequent use'' logit and the binary adoption logit are also non-significant.
For non-text AI, the ordered-logit coefficient is roughly four times larger in
magnitude and highly significant ($\hat\beta = -0.377$,
$\mathrm{SE} = 0.063$, $p < .001$), and the result is consistent across the
ordered, midpoint-binary, and adoption-binary specifications.

\begin{table}[t]
  \centering
  \caption{Demographic-adjusted text vs.\ non-text decomposition. Coefficient
    on a +1 standard-deviation increase in AI literacy, controlling for age,
    income, general knowledge, motivation for autonomy, and male gender.
    Item-level models include task fixed effects.}
  \label{tab:decomposition}
  \begin{tabular}{lrrrl}
    \toprule
    \textbf{Outcome / Model} & \textbf{Coef.} & \textbf{SE} & \textbf{$p$} & \textbf{Notes} \\
    \midrule
    \multicolumn{5}{l}{Text AI only (N$=$401)} \\
    \quad OLS                              & $-0.077$ & $0.082$ & $.351$ & HC3 SE       \\
    \quad Binary logit ($Y > 3$)           & $-0.103$ & $0.141$ & $.467$ & OR $= 0.90$ \\
    \quad Binary adoption ($Y > 1$)        & $-0.061$ & $0.128$ & $.634$ & OR $= 0.94$ \\
    \quad Ordered logit                    & $-0.090$ & $0.104$ & $.387$ & Prop.\ odds \\
    \midrule
    \multicolumn{5}{l}{Non-text AI (4 tools; 1{,}604 item-level obs.)} \\
    \quad OLS on participant avg.\         & $-0.207$ & $0.054$ & $< .001$ & HC3 SE      \\
    \quad Binary logit ($Y > 3$)           & $-0.398$ & $0.103$ & $< .001$ & OR $= 0.67$ \\
    \quad Binary adoption ($Y > 1$)        & $-0.388$ & $0.069$ & $< .001$ & OR $= 0.68$ \\
    \quad Ordered logit                    & $-0.377$ & $0.063$ & $< .001$ & Prop.\ odds \\
    \bottomrule
  \end{tabular}
\end{table}

The contrast is even sharper at the adoption margin. The binary adoption
specification ($Y > 1$, i.e., ``ever used in the last six months'') gives an
odds ratio of $0.68$ per one-standard-deviation increase in AI literacy for
non-text tools, a clean, easily interpretable result, and an odds ratio
close to unity (OR = 0.94) for text AI.

\begin{figure}[t]
  \centering
  \includegraphics[width=0.78\linewidth]{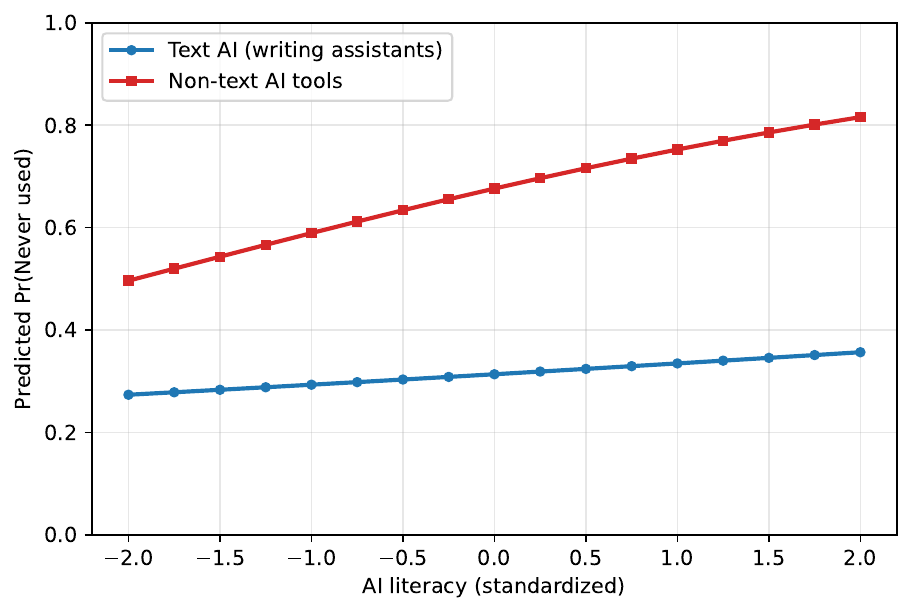}
  \caption{Predicted probability of non-use by AI literacy. Curves
    are derived from ordered-logit fits on text AI only (blue) and on the
    four non-text AI tools (red, with task fixed effects), holding all
    covariates at their sample means. Higher AI literacy predicts much higher
    odds of having never used non-text AI tools; the relationship is roughly
    flat for text AI.}
  \label{fig:predicted}
\end{figure}

Figure~\ref{fig:predicted} visualizes the consequence of this asymmetry. We
plot the ordered-logit-predicted probability $\Pr(Y = 1)$ (``Never used'')
over the empirical range of standardized literacy, separately for the text
and non-text models. For text AI, the predicted probability of non-use rises
only modestly, from $0.27$ at $z = -2$ to $0.35$ at $z = +2$. For non-text
AI, it rises sharply, from $0.50$ to $0.82$ across the same range. In other
words, the higher a respondent's AI literacy, the more likely they are to
have never tried image generators, productivity bots, website builders, or
health apps, while their probability of having used a ChatGPT-style writing
assistant is essentially flat.

\subsection{Original Study~3 control check}

Table~\ref{tab:robustness} repeats the decomposition using the same controls
used for Tully et al.'s original Study~3 regression results: technology
readiness, general knowledge, motivation for autonomy, and male gender. These
are the results reported in Table~4 of \citet{tully2025lower}. Because
technology readiness has missing observations, this specification uses
N$=$379. The non-text pattern strengthens: the ordered-logit coefficient is
$-0.502$ and the adoption odds ratio is $0.61$. The text AI coefficient also
becomes negative and statistically significant in the ordered and
midpoint-binary models, although not at the adoption margin. We therefore avoid
interpreting the text AI result as a true zero. That the text AI estimate crosses 
conventional significance under one covariate set but not the other is 
itself informative: specification sensitivity of this kind is absent for the 
non-text effect, which is large and consistent throughout. The more stable conclusion is
comparative: the literacy gradient is larger and more consistently
adoption-based for non-text AI tools.

\begin{table}[t]
  \centering
  \caption{Robustness check using Tully et al.'s original Study~3 controls.
    Coefficient on a +1 standard-deviation increase in AI literacy, controlling
    for technology readiness, general knowledge, motivation for autonomy, and
    male gender. Item-level models include task fixed effects.}
  \label{tab:robustness}
  \begin{tabular}{lrrrl}
    \toprule
    \textbf{Outcome / Model} & \textbf{Coef.} & \textbf{SE} & \textbf{$p$} & \textbf{Notes} \\
    \midrule
    \multicolumn{5}{l}{Text AI only (N$=$379)} \\
    \quad Binary logit ($Y > 3$)           & $-0.322$ & $0.154$ & $.037$ & OR $= 0.72$ \\
    \quad Binary adoption ($Y > 1$)        & $-0.238$ & $0.145$ & $.100$ & OR $= 0.79$ \\
    \quad Ordered logit                    & $-0.290$ & $0.112$ & $.010$ & Prop.\ odds \\
    \midrule
    \multicolumn{5}{l}{Non-text AI (4 tools; 1{,}516 item-level obs.)} \\
    \quad Binary logit ($Y > 3$)           & $-0.577$ & $0.112$ & $< .001$ & OR $= 0.56$ \\
    \quad Binary adoption ($Y > 1$)        & $-0.497$ & $0.075$ & $< .001$ & OR $= 0.61$ \\
    \quad Ordered logit                    & $-0.502$ & $0.067$ & $< .001$ & Prop.\ odds \\
    \bottomrule
  \end{tabular}
\end{table}

\section{Discussion}

The original aggregate association reported by \citet{tully2025lower} is
robust in a narrow statistical sense: it survives every reasonable parametric
alternative to OLS on an averaged Likert index. But the decomposition shows
that the aggregate coefficient is not clean evidence for a general
lower-literacy, higher-AI-receptivity tendency. The pooled estimate combines three
analytically distinct constructs:

\begin{enumerate}
  \item General AI receptivity: a latent willingness to adopt AI
    across product categories.
  \item Text AI usage intensity: how often a respondent uses
    text-based generative AI tools such as ChatGPT, conditional on having
    adopted them.
  \item Non-text AI adoption breadth: the probability of having
    tried at least one of several heterogeneous AI-labelled consumer
    products.
\end{enumerate}

Our results suggest that Study~3 speaks most clearly to the third construct,
not to general AI receptivity. This is precisely why the text AI result is
important: text AI is the only tool category in Study~3 with all five response
levels well populated, and therefore the cleanest item-level test of a
frequency-based receptivity claim. Yet this margin is weaker and less stable
than the non-text margin: it is small and non-significant in the full-sample
demographic-adjusted specification, but becomes negative and statistically
significant under Tully et al.'s original Study~3 control specification. By
contrast, the non-text relationship remains large across specifications and
appears at both the ordered-response and adoption margins. The aggregate
estimate is therefore carrying substantial information about adoption breadth
across categories where the modal respondent has never used the product.

This re-interpretation is a substantive constraint on the broader theoretical
claim that lower-literacy consumers may perceive AI as magical
\citep{tully2025lower}. At least in the usage dataset, the strongest evidence
is not that lower-literacy consumers are more receptive to AI across the board,
but that they are more likely to have tried several lower-penetration AI-labelled
tools. This also changes what the claim implies for managers. The original
article suggests that firms may benefit from targeting lower-AI-literacy
consumers. Our results narrow this prescription in two ways.
First, the targeting case is much stronger for low-penetration, non-text AI product categories
(image generation, productivity automation, AI design services, AI
wellness apps) than for more familiar text assistants. Second, the
non-text effect is driven primarily by the adoption margin rather than
by intensive use: low-literacy consumers are differentially more likely
to try these products, not necessarily to use them more
intensively conditional on trial. This matters for product strategy.
Lifetime-value calculations that assume a uniform usage-intensity
response to literacy may overestimate the value of low-literacy
acquisitions.

Methodologically, our reanalysis illustrates the value of treating
heterogeneous ordinal items as what they are. Averaging five items with
sharply different empirical distributions into a single ``receptivity''
index is convenient, and in this case it does not invert any sign, but
it does flatten a multi-tool adoption pattern into something that reads
like a single dispositional trait. The construct-validity question that
follows, whether ``AI receptivity'' is one thing or several, deserves
attention beyond the present application
\citep{long2020aileiteracy,ng2021conceptualizing}.

\section{Limitations}

Several limitations of our reanalysis warrant explicit mention. First, we rely
entirely on the data and codebook released by the original authors. Their
Studies~1, 2, and 4--7 use different measures of receptivity, including
cross-country adoption readiness, propensity to use generative AI for
assignments, and relative preference for AI versus human task completion. Our
critique therefore applies most directly to Study~3.

Second, like the original Study~3, our analyses are correlational and rely on
self-reported usage. Any measurement error in self-reported usage that
correlates with literacy could bias the estimated coefficients. The
decomposition itself is also post hoc rather than preregistered, although we
implemented the four model families exactly as described in our analytic
strategy and provide all code.

Third, we make no claim that AI literacy causes reduced adoption of any tool.
The mediation analysis in the original paper, centered on perceptions of AI as
magical, is not addressed here and remains the authors' primary theoretical
mechanism. An obvious extension would be to ask whether perceived
``magicalness'' varies systematically across non-text and text tool categories,
and whether that variation can account for the asymmetry we document. This
would require new data.

Finally, OLS and binary-logit standard errors are heteroskedasticity-robust,
but the ordered-logit estimates use the standard maximum-likelihood covariance
matrix available in \texttt{statsmodels}. A mixed-effects or cluster-robust
ordinal model would be a useful additional robustness check.

\section{Conclusion}

We do not dispute the aggregate lower-literacy/higher-usage association
reported by \citet{tully2025lower}. We do dispute the interpretation that this
association, in Study~3, provides clean evidence for general AI receptivity. When
usage is decomposed by tool type and modelled as ordinal adoption data, the
negative association is stronger and more stable for non-text AI categories,
where it operates primarily at the adoption margin rather than at the intensity
margin. The concern is amplified by the sampling pattern: the only item with
well-populated response categories across the full frequency scale is text AI,
and that item provides the weakest and most specification-sensitive evidence.
Study~3 is therefore better read as evidence about non-text AI adoption breadth,
with weaker evidence about text AI use. The central implication is sharper than
a robustness qualification: the only study in the original article that observes
past usage of real AI tools does not support the broadest version of the
lower-literacy/higher-receptivity claim.

\section*{Data and code availability}

The Study~3 data are publicly available from the original authors at
\href{https://researchbox.org/1491}{ResearchBox \#1491}. All Python code used
to produce the tables and figures in this paper, together with a fully executed
Jupyter notebook, is available in the project repository:
\href{https://github.com/HristoInouzhe/AI-use-vs-AI-literacy}{github.com/HristoInouzhe/AI-use-vs-AI-literacy}.

\bibliography{references}

\end{document}